\documentclass[conference]{IEEEtran}
\IEEEoverridecommandlockouts
\usepackage[utf8]{inputenc}
\usepackage[utf8]{inputenc}
\usepackage[english]{babel}
\usepackage{amsthm}
\usepackage[linesnumbered,ruled,vlined]{algorithm2e}
\usepackage{algpseudocode}
\algrenewcommand\algorithmicrequire{\textbf{Input:}}
\algrenewcommand\algorithmicensure{\textbf{Output:}}
\usepackage{threeparttable}
\usepackage{makecell}
\usepackage{amsmath,amssymb,amsfonts}
\usepackage{amsthm}
{
\theoremstyle{plain}

}
\usepackage{amsmath}
\usepackage{cite}
\usepackage{graphicx}
\usepackage{bbm}
\SetKwRepeat{Do}{do}{while}
\usepackage[outdir=./]{epstopdf}
\usepackage{textcomp}
\usepackage{xcolor}
\usepackage{authblk}

\usepackage{mathtools}

\usepackage[top=0.7in,bottom=0.55in,left=0.72in,right=0.72in]{geometry}
\usepackage{subfigure}
\renewcommand{\theenumi}{\alph{enumi}}

\SetKwInput{KwInput}{Input}                % Set the Input
\SetKwInput{KwOutput}{Output}              % set the Output
\title{Multi-Agent Reinforcement Learning Based Coded Computation  for Mobile Ad Hoc Computing}

\author{Baoqian~Wang$^{1}$ \and Junfei Xie$^{2}$ \and Kejie Lu$^{3}$ \and Yan Wan$^{4}$ \and Shengli Fu$^{5}$
\thanks{Baoqian Wang is with the Department of Electrical and Computer Engineering, University of California, San Diego, and  San Diego State University, San Diego, CA, 92182 (e-mail: {\tt\small bawang@ucsd.edu)}.}
\thanks{Junfei Xie is with the Department of Electrical and Computer Engineering, San Diego State University, San Diego, CA, 92182 (e-mail: {\tt\small jxie4@sdsu.edu}). Corresponding author.} 
	\thanks{Kejie Lu is with the Department of Computer Science and Engineering, University of Puerto Rico at Mayag\"{u}ez, Mayag\"{u}ez, Puerto Rico, 00681, e-mail:  ({\tt\small kejie.lu@upr.edu}).	}
\thanks{Yan Wan is with the Department of Electrical Engineering, University of Texas at Arlington, Arlington, Texas, 76019 (e-mail:  {\tt\small yan.wan@uta.edu}).}
\thanks{Shengli Fu is with the Department of Electrical Engineering, University of North Texas, Denton, Texas, 76201 (e-mail: {\tt\small Shengli.Fu@unt.edu}).}
}

\newcommand{\calB}{{\cal B}}

\newcommand{\bfa}{\mathbf{a}}

\newcommand{\bfd}{\mathbf{d}}

\newcommand{\bff}{\mathbf{f}}

\newcommand{\bfp}{\mathbf{p}}

\newcommand{\bfr}{\mathbf{r}}
\newcommand{\bfs}{\mathbf{s}}

\newcommand{\bfv}{\mathbf{v}}

\newcommand{\bfx}{\mathbf{x}}
\newcommand{\bfy}{\mathbf{y}}

\newcommand{\bfpi}{\boldsymbol{\pi}}

\newcommand{\bfA}{\mathbf{A}}

\newcommand{\bfG}{\mathbf{G}}

\newcommand{\bfX}{\mathbf{X}}

\newcommand{\prl}[1]{\left(#1\right)}

\newcommand{\crl}[1]{\left\{#1\right\}}
\begin{document}

\maketitle
\begin{abstract}
Mobile ad hoc computing (MAHC), which allows mobile devices to directly share their computing resources, is a promising solution to address the growing demands for computing resources required by mobile devices. However, offloading a computation task from a mobile device to other mobile devices is a challenging task due to frequent topology changes and link failures because of node mobility, unstable and unknown communication environments, and the heterogeneous nature of these devices. To address these challenges, in this paper, we introduce a novel coded computation scheme based on multi-agent reinforcement learning (MARL), which has many promising features such as adaptability to network changes, high efficiency and robustness to uncertain system disturbances, consideration of node heterogeneity, and decentralized load allocation. Comprehensive simulation studies demonstrate that the proposed approach can outperform state-of-the-art distributed computing schemes. 
\end{abstract}

\section{Introduction}
In recent years, we have witnessed the significant growth of mobile applications that are based on advanced artificial intelligence and deep-learning models, such as face recognition, %\cite{masi2018deep}, 
object detection % \cite{hu2018relation} 
and natural language processing. %\cite{gardner2018allennlp}, 
These applications stimulate the demands for greater computing resources, which may be difficult to support in one device. To address this need, one promising solution is to offload computation-intensive tasks from resource-limited mobile devices to remote clouds or nearby edge computing facilities \cite{wang2017survey}. Nevertheless, this solution can suffer from high transmission delays and becomes infeasible when there are no communication infrastructures nearby, e.g., in rural areas or emergencies. Local computing is thus essential for delay-sensitive applications, or when there is no or weak wireless Internet connection to remote clouds and when nearby edge computing devices are not available. 

Mobile ad hoc computing (MAHC) \cite{cao2017towards,yaqoob2016mobile} is coined recently to provide local computing services, by enabling resource sharing among mobile devices. However, the research on MAHC is still in its infancy.  Existing studies have mainly considered smartphones \cite{li2015heuristics} or ground vehicles \cite{liu2020vehicular} as the main  resources. A few recent studies attempted to move computing resources up to the unmanned aerial vehicles (UAVs) and create a networked airborne computing system \cite{wang2020computing, wang2018enabling}  that allows UAVs to share resources between each other through direct flight-to-flight communications. This can be considered as one type of MAHC, but studies along this direction are still very limited. 

Offloading a computation task from a mobile device to other mobile devices nearby is challenging, due to the mobility and heterogeneity  of these devices. Particularly,  devices' movement can cause frequent topology changes and communication instability/failures. %Besides, devices may have different computing powers, which can vary over time. 
Furthermore, mobile devices may not have an explicit knowledge of its communication environment and the computing powers of other devices. In this paper, we aim to address these challenges, by exploring coded computation \cite{lee17coded} and multi-agent reinforcement learning (MARL) \cite{zhang2019multi}. 

Coded computation is a promising technique for improving the resilience of distributed computing systems to uncertain system disturbances, such as communication bottlenecks, node/link failures, slow-downs of computing nodes, etc. Different coded computation tasks have been explored for systems with homogeneous computing nodes, such as matrix-vector multiplication \cite{lee18speeding}, gradient descent \cite{tandon17gradient}, multivariate polynomials \cite{ rudow2020locality}, etc. Coded computation for heterogeneous systems has also been recently investigated. For example, \cite{reisizadeh17coded} introduced a \emph{Heterogeneous Coded Matrix Multiplication} (HCMM) scheme  that  maximizes the expected computing results received by the master node. Our research group developed a batch processing based coded computation (BPCC) scheme \cite{wang2019batch, wang2019coding}, which allows partial results to be returned and is highly efficient and robust to uncertain system disturbances. 
However, all these works assume static networks and adopt simplified communication and computing models. Recently, a coded cooperative computation protocol (C3P) \cite{keshtkarjahromi2018dynamic} was proposed that considers mobile computing systems, and allocates tasks in a dynamic and adaptive manner based on the time lag between two consecutive tasks. However, this method requires explicit knowledge of the delay model. 

MARL has been widely used for applications that involve interactions among multiple agents, such as multi-robot control, multiplayer games, Internet of Things, to name a few. Although multiple studies  apply RL/MARL to solve computation offloading or resource allocation problems, e.g., \cite{lu2020optimization, chen2020artificial}, %Moreover, a recent study uses MARL for task offloading for MEC \cite{cao2020multi}. 
the use of MARL for MAHC, especially with coded computation, has not  been investigated. % yet, to the best of our knowledge. 

In this paper, we consider distributed computing over a MAHC system that consists of multiple mobile and heterogeneous computing devices, with unknown computing power and communication environment. As a preliminary investigation, we focus on a classical distributed computing task, the matrix-vector multiplication problem, which is a  building block of many computation tasks and machine learning algorithms. Based on this task, our \emph{main contribution} is an innovative MARL-based coded computation framework with the following key features: 1) adaptability to time-varying communication changes caused by node mobility; 2) capability to address node heterogeneity; 3) applicability for any MAHC or mobile edge/fog computing (MEC) systems without having to know their % the need to construct 
communication or computing characteristics; 
4) high computational efficiency by applying the multi-agent deep deterministic policy gradient (MADDPG) \cite{lowe2017multi}, the state-of-the-art MARL algorithm, and adopting the batch processing procedure proposed in our previous studies \cite{wang2019batch,wang2019coding}; and 5) decentralized load allocation that relieves the computation burden at the master node. To demonstrate these promising features, we conduct simulation studies on a networked airborne computing system.

In the rest of this paper,  Sec. \ref{sec:pro} describes the computing system, distributed computing schemes, and the optimization problem.  Sec. \ref{sec:main_results} transforms the original problem into a MARL problem and then presents a MADDPG- and batch processing based solution. Simulation  results are presented in Sec. \ref{sec:simulation}, followed by the conclusion in Sec. \ref{sec:conclusion}. 

\section{Problem Description}
\label{sec:pro}
In this section, we first describe the computing system and the computation tasks considered in this study, and then present the uncoded and coded distributed computing schemes. Finally, we formulate the optimization problem.

\subsection{Computing System}
Consider a MAHC system that consists of multiple mobile devices with different computing powers. These devices can talk to each other through wireless communication. Suppose one of the devices receives/has a sequence of $K$ matrix-vector multiplication tasks to complete, denoted as $\{\bfA\bfx_1, \bfA\bfx_2,\dots,\bfA\bfx_K\}$, where $\bfA\in\mathbbm{R}^{p\times m}$ is a pre-stored matrix and $\bfx_j \in\mathbbm{R}^{m\times 1}$, $j\in [K]:= \{1,2,\dots,K\}$ are input vectors. %Here we define $[n] = \{1,2,\ldots,n\}$ for any positive integer $n \in \mathbbm{Z}^+$. 
Due to limited computing power, this device, often called the \emph{master node}, offloads the computation tasks to its $N$ neighbors, known as the \emph{worker nodes},  in a sequential order, %Suppose the computation tasks are offloaded in a sequential order, 
with task $j$ being offloaded only after the previous task $j-1$ has been completed. 

Offloading a computation task $j$ is achieved by first partitioning the task into $N$ subtasks.  
%To offload a computation task $j$, the master node first partitions the task into $N$ subtasks and then assigns these subtasks to the $N$ worker nodes. 
Each worker node then executes one subtask and returns the result to the master node after completion. The master recovers and outputs the value of task $j$, i.e., $\bfA\bfx_j$, after receiving sufficient results and then moves on to the next task. In the following subsection, we describe how the traditional uncoded and coded computation schemes partition and allocate a task.

\subsection{Distributed Computing Schemes}
\subsubsection{Uncoded Scheme}
The traditional uncoded scheme partitions a matrix-vector multiplication task $\bfA\bfx_j$ by decomposing matrix $\bfA$ row-wise into $N$ non-overlapping submatrices  $\{\bfA_{1,j}, \bfA_{2,j}, \dots, \bfA_{N,j}\}$, where $\bfA_{i,j} \in \mathbb{R}^{\ell_{i,j}\times m}$. Each worker node $i$ computes subtask $\bfA_{i,j}\bfx_j$, $i\in [N]$. The master node is able to recover $\bfA\bfx_j$ after receiving results from all the worker nodes, by simply concatenating the results, i.e., $\bfA\bfx_j = [\bfA_{1,j}\bfx_j;\bfA_{2,j}\bfx_j;\dots;\bfA_{N,j}\bfx_j]$.

One major drawback of this scheme is that the master node cannot recover $\bfA\bfx_j$ until all results from the worker nodes have been received. Because of this, the efficiency of the uncoded scheme is bounded by the slowest worker node, making it inefficient when uncertain system disturbances are prominent, e.g., in a MAHC system.  

\subsubsection{Coded Scheme}
The coded scheme overcomes the drawback of the uncoded scheme by 
%any straggler in computation nodes will affect the overall performance. To address the issue, we present the coded computation scheme, which 
introducing redundant computations using the coding theory. In particular, for each task $j$, the coded scheme first encodes $\bfA$ into a larger matrix ${\hat{\bfA}}_j\in \mathbb{R}^{q_j\times m} $ with additional rows, i.e., $q_j > p $, by applying the equation below
\begin{equation}
\label{eq:encode}
\hat{\bfA}_j=\bfG_j \bfA.  
\end{equation}
where $\bfG_j \in \mathbb{R}^{q_j\times p}$ is an encoding matrix with the property that any matrix formed by any $p$ rows of $\bfG_j$ has full rank.  %vectors are linearly independent from each other \cite{lee17coded}. In other words, we can use any $p$ rows of $\bfG_j$ to create an $p \times p$ full-rank matrix. 
With $\hat{\bfA}_j$, the coded scheme then follows a similar procedure to partition and allocate the task. Particularly, worker node $i$ executes subtask $\hat{\bfA}_{i,j}\bfx_j$, where $\hat{\bfA}_{i,j} \in \mathbb{R}^{\ell_{i,j}\times m}$, $i\in N$ is a submatrix of $\hat{\bfA}_j$, and $\sum_{i=1}^N \ell_{i,j} = q_j$. 

To recover the final result $\bfA\bfx_j$, the master node waits until sufficient, but not all, results are received. Specifically, let $\hat{\bfy}_{j}$ be the results received by the master node by a certain time, which satisfies  $\hat{\bfy}_{j}=\hat{\bfG}_{j}\bfA \bfx_j$, where $\hat{\bfG}_{j}$ is a submatrix of $\bfG_{j}$. $\bfA\bfx_j$ can then be recovered when $|\hat{\bfy}_j|\geq p$ via:
\begin{equation}
\label{eq:decoding}
\bfA\bfx_j=(\hat{\bfG}_{j}^\top\hat{\bfG}_{j})^{-1}\hat{\bfG}_{j}^\top\hat{\bfy}_{j},
\end{equation}

\subsection{Optimization Problem}
In  distributed computing schemes, the load numbers $\ell_{i,j}$ determine the amount of workloads assigned to each worker node. As worker nodes have different computing powers and the time required by each node to transmit the result is also different and time-varying, it is crucial to choose a proper load number $\ell_{i,j}$ for each worker node $i\in [N]$ and each task $j\in[K]$. To determine the optimal load numbers $\ell^*_{i,j}$, we formulate an optimization problem to minimize the expected task completion time. 

Denote the time required by worker node $i$ to complete the $j$-th task, i.e., $\hat{\bfA}_{i,j}\bfx_j$ with $\hat{\bfA}_{i,j}\in\mathbb{R}^{\ell_{i,j}\times m}$, as $T_{i,j}$. Then $T_{i,j}$ can be captured by the following equation:
\begin{equation}
    T_{i, j} \approx T_{i, comm}(\bfx_j) + T_{i, comp}(\hat{\bfA}_{i,j},\bfx_j) + T_{i, comm}(\hat{\bfA}_{i,j}\bfx_j)
\end{equation}
where $T_{i, comm}(\bfx_j)$ is the time spent to send $\bfx_j$ from the master node to the worker node $i$, 
%of receiving $\bfx_j$ from master node, 
$T_{i, comp}(\hat{\bfA}_{i,j},\bfx_j)$ is the time spent to compute subtask $\hat{\bfA}_{i,j}\bfx_j$, and $T_{i, comm}(\hat{\bfA}_{i,j}\bfx_j)$ is the time spent to send the computation result back to the master node. Note that $T_{i,j}$ is a random variable reflective of various environmental and system uncertainties. Explicit communication and computation models are not available to the nodes. 

With $T_{i,j}$, we can then describe the task completion time for each task $j \in [K]$ as follows: 
%The completion time for each task $j$ denoted by $T_j$ is then described as
\begin{equation}
    T_j = \underset{t}{\min} \ \ \{t ~~| \ \ R_{j}(t) \geq p\}   
\end{equation}
where $R_{j}(t) = \sum_{i=1}^{N}\ell_{i,j}\mathbbm{1}_{T_{i, j}\leq t}$ is the total number of rows of inner product results for task $j$ that the master node has received by time $t$. $\mathbbm{1}$ is the indicator function \cite{kenny2003indicator}.
Finally, the optimization problem to solve can be formulated as follows,
\begin{align} 
\label{eq:main_problem}
& \underset{\ell_{i,j}, \forall i\in[N], \forall j \in[K]} {\text { minimize }} 
& ~
&\mathbb{E}\left[\sum_{j=1}^{K}T_j\right] \\
& \text { subject to } 
& 
& \ell_{i,j}\in  \mathbb{Z}^+, \forall i \in [N], \forall j \in [K]\nonumber\\
& 
& 
& \sum_{i=1}^{N}\ell_{i,j} \geq p, \forall i \in [N], \forall j \in [K] \label{eq:constranit}
%& p_i \leq \ell_i, p_i \in \mathbb{Z}^+,\forall i \in [N]
\end{align}

\section{Main Results}
\label{sec:main_results}
In this section, we solve the optimization problem in \eqref{eq:main_problem}, by first transforming it into a MARL problem and then developing a MADDPG- and batch processing based solution.

\subsection{MARL-based Formulation}
\label{sec:marl}
To solve the optimization problem in \eqref{eq:main_problem}, we provide a MARL-based formulation. In particular, the
%By using MARL, 
$N$ worker nodes are regarded as the \emph{agents} that learn to achieve the optimization goal in \eqref{eq:main_problem} by interacting with the environment. % to receive reward signals.  
The \emph{state} of each agent $i$ at time $t$ is denoted as $\bfs_{i}(t) = [d_i(t), \bfd_{-i}(t), \bfv_i(t), \bfv_{-i}(t), \bfv_{m}(t)]^\top\in \mathcal{S}_i$, where $d_i(t)$ and $\bfd_{-i}(t)$ represent the distance of agent $i$ to the master node and the distances of all agents except agent $i$ to the master node at time $t$, respectively. $\bfv_{i}(t)$,  $\bfv_{-i}(t)$, and $\bfv_m(t)$ represent the velocity of agent $i$, velocities of all agents except agent $i$, and  velocity of the master node at time $t$, respectively. $\mathcal{S}_i$ is the corresponding state space. 

Upon start of a new task $j$, each agent $i$ decides the  computation load it will execute for this task. This differs from the traditional load allocation mechanisms where the master node decides the computation load for each worker node (agent). Now denote the time when the master node sends task $j$ to the agents as $t_j$, we then define the
\emph{action} of each agent $i$ taken at time $t_j$ as $a_i(t_j) = \ell_{i,j} \in \mathcal{A}_i$, where $\mathcal{A}_i$ is the corresponding action space. It is reasonable to set $\mathcal{A}_i = [0, p]$, as the total  computation load required for completing task $j$ is $p$. %  task size need to be completed is $\bfA x_j$ with $\bfA\in\mathbb{R}^{p\times m}$. 
Note that, during the execution of task $j$, agent $i$ does not take any actions, and the next action is taken when the next task ${j+1}$ starts.  The transition of the agent's state from $\bfs_{i}(t_j)$ to  $\bfs_i(t_{j+1})$ is determined by the mobility model of the agent described by $\bfs_i(t_{j+1}) = f_i(\bfs_i(t_j))$, and $t_{j+1}= t_j + T_j$, where $T_j$ is the time spent to complete task $j$. Here we assume the velocities of all agents and the master node do not change during the execution each task $j$.

After each transition, agent $i$ receives a reward $r_i(\bfs(t_j), \bfa(t_j)): \mathcal{S} \times \mathcal{A} \mapsto \mathbb{R}$, where $\bfs(t) = (\bfs_1(t), \bfs_2(t),\ldots,\bfs_N(t)) \in \mathcal{S}: = \prod_{i \in [N]} \mathcal{S}_{i}$ and $\bfa(t)  = (a_1(t), a_2(t),\ldots,a_N(t)) \in \mathcal{A}: =\prod_{i \in [N]} \mathcal{A}_{i}$ 
are the joint state and action for all agents, respectively, and $\mathcal{S}$, $\mathcal{A}$ are the corresponding joint state and action spaces. Here the reward function is defined as $r_i(\bfs(t_j), \bfa(t_j))=-T_j - c\mathbbm{1}_{\sum_i^{N}\ell_{i,j}\leq p}$, which takes the constraint \eqref{eq:constranit} into  consideration and $c$ is the weight for the penalty term. 
Given state $\bfs_i$, each agent then aims to choose a deterministic policy, specified by function $\pi_i(\bfs_i): \mathcal{S}_i \mapsto \mathcal{A}_i$, % that outputs agent's $a_i\in\calA_i$, 
such that the expected cumulative discount reward given by 
 \begin{align}
\label{eq:value_function}
V^{\bfpi}_i(\bfs) := \mathbb{E}_{\substack{\bfs(t_j)\sim \bff\\ \bfa(t_j)\sim \bfpi}} \biggl[\sum_{j=1}^{K}\gamma^{j-1} r_i(\bfs(t_j),\bfa(t_j)) |\nonumber\bfs(t_1)=\bfs\biggr]
\end{align}
is maximized. In the above equation, $\gamma$ is a discount factor, $\bfpi = (\pi_1, \pi_2, \ldots, \pi_N)$ is the concatenated policy of all agents, and $\bff=(f_1, f_2,\ldots,f_N)$ is the concatenated mobility models of all agents. $V^{\bfpi}_i(\bfs)$ is also known as the value function. Alternatively, an optimal policy $\pi_i^*$ for agent $i$ can be obtained by maximizing the action-value function:
\begin{align}
Q_i^{\bfpi}(\bfs,\bfa) := &\mathbb{E}_{\substack{\bfs(t_j)\sim\bff\\ \bfa(t_j)\sim \bfpi}}\biggl[\sum_{j=1}^{K}\gamma^{j-1} r_i(\bfs(t_j),\bfa(t_j))| \nonumber\\&\bfs(t_1)=\bfs, \bfa(t_1)=\bfa\biggr]
\end{align}
and setting $\pi_i^*(a_i | \bfs_i) \in  \arg\max_{a_i} \max_{\bfa_{-i}} Q_i^{*}(\bfs,\bfa)$, where $Q_i^{*}(\bfs,\bfa) := \max_{\bfpi} Q_i^{\bfpi}(\bfs,\bfa)$ and $\bfa_{-i}$ denotes the actions of all agents except $i$.

 \subsection{MADDPG- and Batch-Processing based Solution}
 \label{sec:method}
 To solve the above MARL problem, we apply an advanced  MARL algorithm proposed recently, called multi-agent deep deterministic policy gradient (MADDPG) \cite{lowe2017multi}. 
 The MADDPG is an off-policy algorithm that extends the deep deterministic policy gradient (DDPG) method to multiple agents. 
% Unlike the general formulation for MARL described in Section~\ref{sec:marl}, MADDPG adopts a deterministic policy $\pi_i(s_i)$ for each agent $i \in [N]$, which only depends on the local state $s_i$. 
%The value function $Q_i^{\boldsymbol \pi}(\bfs,\bfa)$ of agent $i$ still depends on the joint state $\boldsymbol s$ and joint action $\bfa$. Moreover, 
For each agent $i$, it uses
    four neural networks to approximate its policy $\pi_i(\bfs_i ; \theta_{\pi_i})$, action-value function $Q_i^{\bfpi}(\bfs, \bfa; \theta_{Q_i})$, target policy $\pi_i'(\bfs_i; \theta_{\pi_i}')$, and target action-value function $Q_i^{\bfpi'}(\bfs, \bfa; \theta_{Q_i}')$, respectively, where $\bfpi' = (\pi'_1, \pi'_2, \ldots, \pi'_N)$ is the concatenated target policy of all agents. A replay memory denoted by $\mathcal{D}$  is used to store the transitions $\crl{(\bfs, \bfa, \bfr, \bfs')}$, where %$\bfs\in\mathcal{S},\bfa\in\mathcal{A},
    $\bfr=[r_1(\bfs,\bfa), r_2(\bfs,\bfa),\ldots,r_N(\bfs,\bfa)]$ and $\bfs'=\bff(\bfs)$. In each training iteration, a mini-batch $\mathcal{B}_i$ is sampled  from the replay memory $\mathcal{D}$ and used to learn the parameters for agent $i$. % $i$'s training.  
In particular, the parameters $\theta_{Q_i}$ for the action-value function  are updated by minimizing the temporal-difference error given as follows:
\begin{align}
\label{eq:maddpg_loss}
& J(\theta_{Q_i}) \nonumber \\ = &\frac{1}{|\calB_i|} \!\sum_{(\bfs,\bfa,\bfs',\bfr)\in \mathcal{B}_i} \!\! \prl{ L_i^{\bfpi'}(\bfs',r_i)\!-\!Q_{i}^{\bfpi}\prl{\bfs, \bfa ; \theta_{Q_i}} }^{2}
\end{align}
where $L_i^{\bfpi'}(\bfs',r_i)=r_i(\bfs, \bfa) + \gamma Q_i^{\bfpi'}(\bfs',\bfpi'(\bfs'))$. The policy parameters $\theta_{\pi_i}$ are updated using gradient ascent based on the policy gradient theorem, with the gradient provided as follows\cite{sutton2018introduction}:
\begin{align}
    \label{eq:maddpg_actor}
    &\nabla_{\theta_{\pi_i}} J(\theta_{\pi_i})\nonumber\\ \approx& \frac{1}{|\calB_i|} \sum_{(\bfs,\bfa,\bfs',\bfr)\in \mathcal{B}_i} \nabla_{\theta_{\pi_i}} \pi_i(\bfs_i;\theta_{\pi_i}) \nabla_{a_i} Q_{i}^{\bfpi}\prl{\bfs, \bfa}.
\raisetag{2ex}
\end{align}

The parameters for the target policy and the target action-value function are updated via Polyak averaging using the following equations:
\begin{equation}
\begin{gathered}
\label{eq:target_update}
\theta^{\prime}_{\pi_i} \leftarrow \tau\theta^{\prime}_{\pi_i}+(1-\tau)\theta_{\pi_i}\\
\theta^{\prime}_{Q_i} \leftarrow \tau\theta^{\prime}_{Q_i} +(1-\tau)\theta_{Q_i},
\end{gathered}
\end{equation}
where $\tau \in (0,1)$ is a hyperparameter. The procedure of MADDPG is summarized in Algorithm \ref{alg:maddpg}.
\begin{algorithm}[h] 
    \DontPrintSemicolon
    \tcp{Initialize parameters}
    Initialize $\theta_{\pi_i}, \theta_{Q_i}, \theta'_{\pi_i}, \theta'_{Q_i}, \forall i \in [N]$\\
    \For{$\text{iteration} = 1:\text{max\_iteration}$}
    {
            \tcp{Collect transitions}
   \For{$k = 1:\text{max\_episode\_number}$}
        {
        \For{$j= 1:K$ }
        {
            For each agent $i$, select $a_{i}(t_j)\!=\! \pi_i(\bfs_i(t_j); \theta_{\pi_i})$.\\
            Execute joint action $\bfa(t_j)$ and receive new state $\bfs'(t_{j+1})$ and reward $\bfr(t_{j+1})$.\\
            Store $\!(\bfs(t_j), \bfa(t_j), \bfr(\bfs(t_j), \bfa(t_j)), \bfs'(t_{j+1})\!)$ in replay buffer $\!\mathcal{D}$.
        }
        }
       %     \tcp{Sample batches from the replay memory}
            Sample a random mini-batch $\mathcal{B}_i$  for each agent $i, \forall i \in [N]$.\\
            
            \tcp{Update parameters }
            \For{$i = 1:N$ }{
            Update $\theta_{Q_i}$  by minimizing the temporal-difference error in  \eqref{eq:maddpg_loss}.\\
            Update $\theta_{\pi_i}$  using gradient ascent, with the gradient provided in \eqref{eq:maddpg_actor}.\\
            
            Update $\theta^{\prime}_{\pi_i}$ and $\theta^{\prime}_{Q_i}$ using  \eqref{eq:target_update}.
            }
    }
\caption{MADDPG Algorithm} \label{alg:maddpg}
\end{algorithm}

To further improve the computational efficiency, we apply the batch processing procedure developed in our previous studies \cite{wang2019batch,wang2019coding}. In particular, we let each worker node $i$ further divide matrix $\hat{\bfA}_{i,j}$ into a set of submatrices $\{\hat{\bfA}_{i,j,k}\}_{k=1}^{w_{i,j}}$ row-wise, namely batches, where each submatrix $\hat{\bfA}_{i,j,k}$ has $b_{i,j}$ number of rows and  $w_{i,j}=\lceil\frac{\ell_{i,j}}{b_{i,j}}\rceil$ is the total number of batches. Note that the last batch has a size of $\ell_{i,j}-(w_{i,j}-1)b_{i,j}$. The worker node then multiples the input vector $\bfx_j$ with each batch one by one and sends the partial result back to the master once available. In \cite{wang2019batch,wang2019coding}, we have shown  nice properties of this batch processing procedure in improving system's efficiency and the resilience to uncertain system disturbances through both theoretical and experimental studies. 
Algorithm \ref{alg:master} summarizes the procedures followed by the master and worker nodes during the execution of computation tasks. 

\begin{algorithm}[h] 
    \DontPrintSemicolon
    \KwInput{$ \bfx_1, \bfx_2, \dots, \bfx_K$}
    \KwOutput{$\bfA \bfx_1, \bfA \bfx_2, \dots, \bfA \bfx_K$}
    \tcp{{Master node}:}
    \For{$j=1:K$}
    {
        $\hat{\bfy}_{j} \leftarrow [~]$  \\
        Broadcast $\bfx_j$ to all worker nodes.\\
                \Do{$|\hat{\bfy}_{j}| < p$}
        {
        Listen to the channel and collect results $\hat{\bfA}_{i,j,k} \bfx_j$ from the worker nodes.\\
  %Continue receiving results\\
            $\hat{\bfy}_{j}  \leftarrow [\hat{\bfy}_{j} ; \hat{\bfA}_{i,j,k} \bfx_j]$}
        Send acknowledgements to all worker nodes.\\
        $\bfA \bfx_j \leftarrow (\hat{\bfG}_{j}^\top\hat{\bfG}_{j})^{-1}\hat{\bfG}_{j}^\top\hat{\bfy}_{j}$  \\
        
          \textbf{return} $\bfA \bfx_j$ %, \forall j \in \{1,2,...,K\}$
        }
        
    \tcp{{Worker node $i$}:}
   % Upon receiving $\bfx_j$: \\
    %\ \ \ \ $\hat{\bfA}\leftarrow \bfG_j \bfA$\\
  \For{$j=1:K$}{
    Determine $\ell_{i,j}$ by applying the policy trained using Algorithm \ref{alg:maddpg}.\\
  \textit{Upon receiving $\bfx_j$}: \\
    Select $\ell_{i,j}$ rows from  $\hat{\bfA}$ to construct $\hat{\bfA}_{i,j}$.\\ 
    Divide $\hat{\bfA}_{i,j}$ into batches $\hat{\bfA}_{i,j,k},k\in[w_{i,j}]$\\
    \For{$k=1:w_{i,j}$}{
    \If{Acknowledgement received}{break}
    Compute $\hat{\bfA}_{i,j,k}\bfx_j$. \\
    Send $\hat{\bfA}_{i,j,k} \bfx_j$ back to the master node.\\}
    }
\caption{Batch Processing Based Coded Scheme} \label{alg:master}
\end{algorithm}

\section{Simulation Studies}
\label{sec:simulation}
In this section, we conduct simulation studies to evaluate the performance of the proposed MADDPG- and batch-processing based algorithm. As an illustration, we consider a MAHC system formed by multiple UAVs. All experiments are run on an Alienware Desktop with 32GB memory, 16-cores CPU with 3.6GHz. In the rest of the section, we first describe the simulation settings and then show the results. 

\subsection{Simulation Settings}
\subsubsection{Environment Description}
We consider a MAHC system that consists of multiple UAVs flying at the same altitude \cite{liu2019learning}. Each UAV is equipped with a micro-computer with different computing powers. The communication between two UAVs is achieved through directional antennas for extended communication range and higher bandwidth. Assume the directional antennas are always aligned during the movement, which can be achieved by the control algorithms developed in  \cite{liu2019learning}. We can then describe the time to transmit a matrix $\bfX \in \mathbb{R}^{m_1 \times n_1}$ between two UAVs as $T_{comm}(\bfX) = \frac{m_1\times n_1 \times u}{C}$,
%\vspace{-0.1cm}
% \begin{equation}
%     T_{comm}(\bfX) = \frac{m_1\times n_1 \times u}{C} 
% \end{equation}
where $u$ (bits) is the average size of the elements in matrix $\bfX$. 
% \begin{equation}
% \label{eq:shanoon}
  $  C = Wlog_2(1+\frac{S}{Noise})$ (bits/sec) is the data rate, 
% \end{equation}
according to the Shannon's theory, where $W$ (Hz) is the communication bandwidth between the two UAVs, $Noise$ (Watts) is the noise power, which is assumed to be constant. $S$ (Watts)
is the signal power that can be modeled by 
% \begin{equation}
%     \begin{aligned}
% \label{eq:rssi}
$S=10^{\frac{(S_d-30)}{10}}$, %\\
% \end{aligned}
% \end{equation}
with $S_d=P_{t}+20 \log _{10}(\lambda)-20 \log _{10}(4 \pi) -20 \log _{10}(d)+G_{l \mid d B i}+\omega$. Here, $P_{t}$ (dBm) is the transmitting power, $\lambda$ is the wave length,  $G_{l \mid d B i}$ is the sum of the transmitting and receiving gains, $\omega$ is Gaussian noise with zero mean and variance $\sigma$, and $d$ is the distance between the two UAVs \cite{liu2019learning}.

To capture the computation time, we adopt the modeling technique used in many recent studies \cite{lee16speeding,reisizadeh17coded}.
Particularly, we assume the time $T_{comp}(\bfA, \bfx)$ to compute a matrix-vector multiplication task $\bfA\bfx$ with $\bfA\in \mathbbm{R}^{\ell\times m}$ and $\bfx \in \mathbbm{R}^{m\times 1}$ follows an exponential distribution defined by:
\begin{equation}
\label{eq:exponential}
\operatorname{P}\left[T_{comp}(\bfA, \bfx)\leq t\right]=1-e^{-\frac{\beta}{\ell}\left(t-\alpha \ell\right)}
\end{equation}
where $t \geq \alpha \ell$. $\beta$ and $\alpha$ are straggling and shift parameters, respectively, and both $\beta$ and $\alpha$ are positive constants. 

To describe the mobility of the UAVs, we here adopt a simple model: 
%The mobility model $\bfs_i(t_{j+1}) = f_i(\bfs_i(t_j))$ is described as
$\bfp(t')=\bfp(t)+\bfv(t)(t'-t)$, 
where $\bfp(t)$ represents the position of a UAV at time $t$, and $t' > t$. In simulations, we assume the velocity $\bfv(t)$ of each UAV is constant, i.e., $\bfv(t) = \bfv$. %The distance between two UAVs, $i$ and $j$, can then be 
\medskip
\subsubsection{Distributed Computing Schemes} 

For comparison, we consider the following three distributed computing schemes as the benchmarks:
\begin{itemize}
\item \textbf{Uniform Uncoded}: This is an uncoded computation scheme which divides the computation loads equally, i.e., $\ell_i = \frac{p}{N}$, $\forall i \in [N]$.

\item \textbf {Load-Balanced Uncoded} \cite{reisizadeh2019coded}: This is an uncoded computation scheme which divides the computation loads according to the computing capabilities of the worker nodes. In particular, the computation load assigned to each worker node $i$ is inversely proportional to the expected time for this node to compute an inner product, i.e.,  $\ell_i \propto (\frac{\beta_i}{\alpha_i \beta_i +1})$ and $\sum_{i=1}^N \ell_i = p$. 

\item \textbf{HCMM} \cite{reisizadeh2019coded}: This is a state-of-the-art coded scheme that computes the load number by %uses the load assignment method in \cite{reisizadeh2019coded}. In particular, 
$\ell_{i} = \frac{p}{h \lambda_{i}}$, where $\lambda_{i}$ is the positive solution to $e^{\beta_i\lambda_{i}}=e^{\alpha_i\beta_i}(\beta_i\lambda_{i}+1)$, $h = \sum_{i=1}^{N}\frac{\beta_i}{1+\beta_i\lambda_{i}}$, and $\beta_i$, $\alpha_i$ are the straggling and shift parameters for worker node $i$, respectively.%Note that this method doesn't consider the mobility of worker nodes either.
\end{itemize}
%Note that all these benchmark schemes neglect the impact of node mobility and are not adaptive to network changes. 
Note that both load-balanced uncoded and HCMM schemes require  explicit knowledge of the nodes' computing capabilities, which is not required by our method.

\subsubsection{Computation Scenarios}
We consider three computation scenarios described as follows:
\begin{itemize}
\item \textbf{Scenario 1:} $N=3$, $p=6000$.
\item \textbf{Scenario 2:} $N=4$, $p=8000$.
\item \textbf{Scenario 3:} $N=5$, $p=10000$.
\end{itemize}
In all scenarios, the total number of tasks is set to $K = 30$, and the size of each task $\bfx_j$ is $m = 10000$. The initial position of each UAV is sampled uniformly from the range $[-100, 100]^2$, and its velocity is randomly selected from the range $[-10m/s, 10m/s]^2$ at each episode. The computing parameter $\beta_i$ of each worker node $i$ is randomly sampled from the range $[10^4, 10^5]$ and $\alpha_i$ is set to  $\alpha_i = \frac{1}{\beta_i}$.  The communication model is configured by setting $W=10^4$, $Noise = 1.1\times10^{-12}$,  
$S_d = 6-20log_{10}(d)$, and $\sigma = 1$.

In our MADDPG- and batch processing based algorithm, the learning rates are set to $\alpha=0.01$ and $\gamma = 0.95$. To approximate the policies and action-value functions, we use neural networks with four fully connected layers. Each hidden layer has $64$ ReLU units. % and ReLU is adopted as the activation function. 
At the output layer, Sigmoid units are used by the policy networks, % so that the outputs are in the range of $[0,1]$. 
and linear units are used by the action-value networks. %The length of each episode $K$ is set to $30$. 
The weight for the penalty term in each reward function is set to $c=200$. 

\subsection{Experimental Results}
In this subsection, we first show the effectiveness of our  algorithm and then investigate the impact of an important parameter, batch size $b_{i,j}$, on the computing performance. Comparison results with the benchmark schemes are  presented at the end.
\subsubsection{Effect of Training}
\begin{figure}[!h]
\centering
		\includegraphics[width=0.3\textwidth]{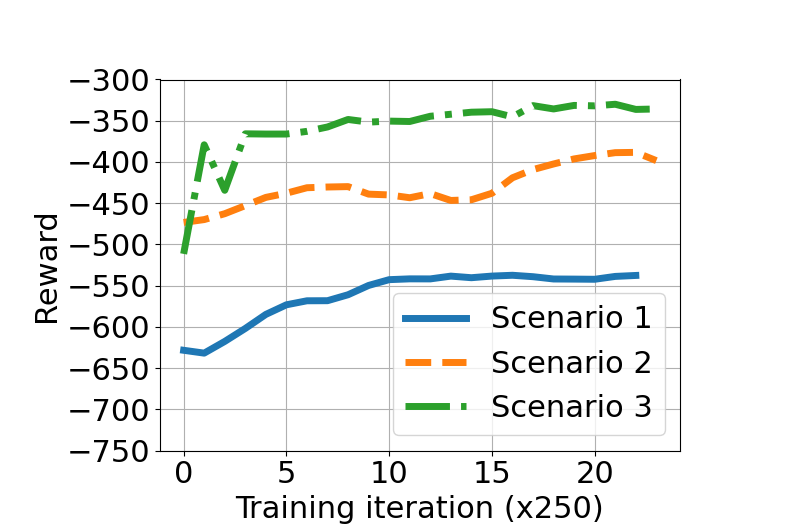}
		\vspace*{-0.3em}
	\caption{Training reward of our algorithm in different scenarios.}
	    \label{fig:training_reward}
\end{figure}

Fig.  \ref{fig:training_reward} shows the training reward of our algorithm in the three scenarios. Each value is obtained by averaging the total rewards of all agents for every  $250$  iterations over $1000$ episodes. We can see that  in all scenarios, the average total reward increases and finally converges with the increasing  numbers of training iterations.

\subsubsection{Impact of  Batch Size}
We investigate the impact of batch size $b_{i,j}$ on the performance of our algorithm, by varying $b_{i,j}$. For each value of $b_{i,j}$, we run Algorithm \ref{alg:master} to measure the task completion times, with load numbers $\ell_{i,j}$ determined by the policies trained using Algorithm \ref{alg:maddpg}. To also understand the resilience of our algorithm to uncertain system disturbances, we further consider the scenario with one uncertain straggler and the scenario without any stragglers. %two straggler settings, i.e., no straggler and 1 straggler, where the straggler effect is simulated by
The straggler is introduced by randomly selecting a worker node at each iteration to sleep for 10 times  of the computation time of that node  using the $time.sleep()$ function.

\begin{figure}[!h]
\centering
\vspace*{-0.8em}
\hspace*{-2.8em}
    \subfigure[]{
		\includegraphics[width=0.28\textwidth]{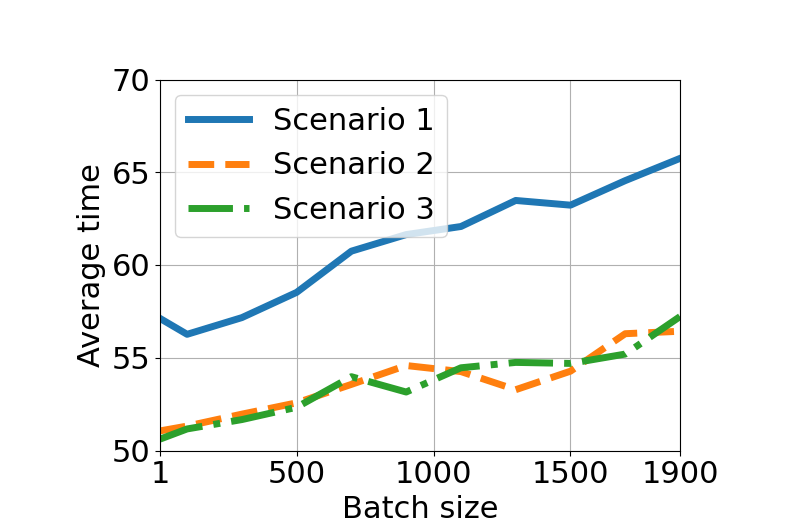}}
		\vspace*{-0.6em}\hspace*{-2.8em}
	\subfigure[]{
	\includegraphics[width=0.28\textwidth]{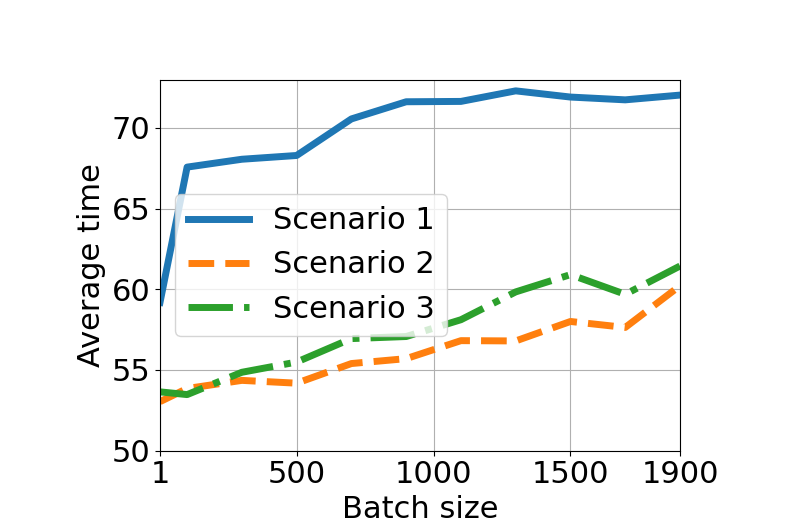}}
	\vspace*{-0.3em}
	\hspace*{-2.8em}
	\caption{The  average total task completion time for different batch sizes when there is (a) no straggler and (b) one uncertain straggler in different scenarios.}
	    \label{fig:batch}
\end{figure}

Fig.  \ref{fig:batch} shows the  total completion time of all $K$ tasks averaged over $20$ episodes in different scenarios. As we can see, with the increase of  batch size, the performance degrades, which is consistent with our findings \cite{wang2019batch}. In the following comparative studies, we set the batch size to $b_{i,j} = 1$, $\forall i\in[N]$ and $j \in [K]$.
%\vspace{-0.7cm}

\subsubsection{Comparative Studies}
We compare our MADDPG- and batch processing based algorithm with the three benchmark schemes in different scenarios. %Similarly, we consider both the scenario when one straggler presents and the scenario without any stragglers. 
For the load-balanced uncoded and HCMM schemes, computing parameters $\beta_i$ and $\alpha_i$ are assumed to be known. 

\begin{figure}[!h]
\centering
\vspace*{-0.4em}
        \subfigure[]{
		\includegraphics[width=0.22\textwidth]{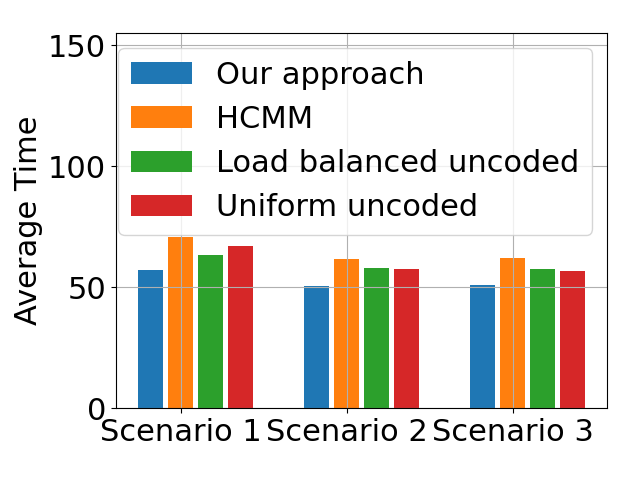}
		 \label{fig:comparison_timea}}
		 \vspace*{-0.6em}\hspace*{-0.6em}
		\subfigure[]{
	\includegraphics[width=0.22\textwidth]{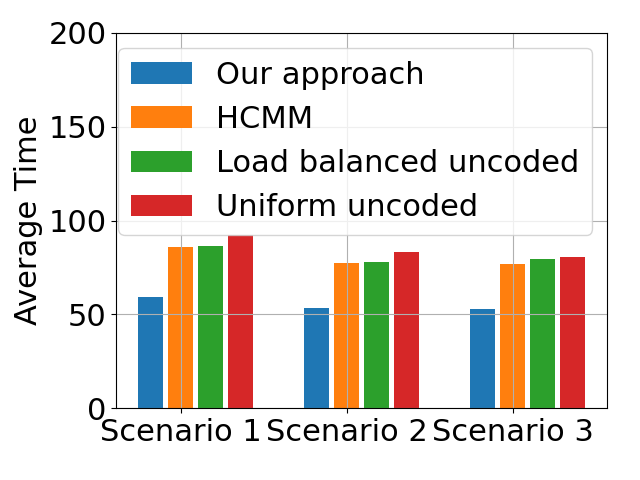}
	 \label{fig:comparison_timeb}}
	 \vspace*{-0.3em}
	\caption{The  average task completion time of each scheme when there is (a) no straggler and (b) one uncertain straggler in different scenarios.}
	    \label{fig:comparison_time}
\end{figure}

 %$0$ straggler setting and $1$ straggler setting for three scenarios. 
The  average total task completion time of each scheme %averaged over $20$ episodes  
in different scenarios is shown in Fig.  \ref{fig:comparison_time}. As we can see, our algorithm achieves the best performance in all scenarios regardless of whether there are any uncertain stragglers. This verifies the effectiveness of our algorithm in addressing node mobility, and also demonstrates its high computational efficiency and resilience to uncertain system disturbances. Of interest, when no stragglers are present (see Fig.  \ref{fig:comparison_timea}), the two uncoded schemes outperform HCMM. This is because HCMM introduces redundant computations, causing additional computation overhead. However, when stragglers are present (see Fig.  \ref{fig:comparison_timeb}), HCMM achieves better performance than the two uncoded schemes, due to the use of the coding technique. Additionally, the uniform uncoded scheme shows the worst performance in most scenarios as it ignores the heterogeneity nature of the computing nodes.

\section{Conclusion}
\label{sec:conclusion}
In this paper, we present a MARL-based coded computation framework for the MAHC system that consists of multiple mobile computing devices sharing resources among themselves. This framework addresses both node mobility and the heterogeneity nature of mobile devices, and can be applied to any MAHC systems without having to know their communication or computing characteristics. %with arbitrary computation and communication characteristics. 
Moreover, our framework, which applies the MADDPG, a state-of-the-art MARL algorithm,  enables the load allocation to be determined in a decentralized manner, thus helping to relieve the computing burden for the master node. It also adopts a batch processing procedure developed in our previous studies to further improve the performance in the aspects of both efficiency and resilience to uncertain system disturbances. The comprehensive simulation studies %in various scenarios with different number of agents, task sizes, number of stragglers 
show that our MADDPG- and batch processing based algorithm achieves the best performance compared with existing uncoded and coded schemes, including the uniform uncoded, load balanced uncoded, and HCMM. %Moreover, we also observe that when there is no straggler. Uniform uncoded and load balanced uncoded outperform HCMM, as HCMM introduce redundant computations, which brings overhead.
In the future work, we will extend this study to consider more general scenarios, such as varying node velocities and task sizes.

\section*{Acknowledgement}
We would like to thank the National Science Foundation (NSF) under Grants CI-1953048/1730675/1730570/1730325, ECCS-1953049/1839804, CAREER-2048266 and CAREER-1714519  for the support of this work.

\begin{small}
	\bibliographystyle{IEEEtran}
	\bibliography{IEEEabrv, reference}
\end{small}

\end{document}